%% file: main.tex
\documentclass{article}

\usepackage{iclr2026_conference,times}
\input{math_commands.tex}

\usepackage{amsmath,amssymb,amsfonts}
\usepackage{booktabs}
\usepackage{graphicx}
\usepackage{subcaption}
\usepackage{hyperref}
\usepackage{url}
\usepackage{enumitem}

\title{Direct Learning of\\Calibration-Aware \mbox{Uncertainty}\\for Neural PDE Surrogates}
\author{
Carlos Stein Brito \\
NightCity Labs, Lisbon, Portugal \\
\texttt{carlos.stein@nightcitylabs.ai}
}
\date{}

\iclrfinalcopy

\begin{document}
\maketitle
\thispagestyle{plain}
\pagestyle{plain}

\begin{abstract}
Neural PDE surrogates are often deployed in data-limited or partially observed regimes where downstream decisions depend on calibrated uncertainty in addition to low prediction error. Existing approaches obtain uncertainty through ensemble replication, fixed stochastic noise such as dropout, or post hoc calibration. Cross-regularized uncertainty learns uncertainty parameters during training using gradients routed through a held-out regularization split. The predictor is optimized on the training split for fit, while low-dimensional uncertainty controls are optimized on the regularization split to reduce train--test mismatch, yielding regime-adaptive uncertainty without per-regime noise tuning. The framework can learn continuous noise levels at the output head, within hidden features, or within operator-specific components such as spectral modes. We instantiate the approach in Fourier Neural Operators and evaluate on APEBench sweeps over observed fraction and training-set size. Across these sweeps, the learned predictive distributions are better calibrated on held-out splits and the resulting uncertainty fields concentrate in high-error regions in one-step spatial diagnostics.
\end{abstract}

\section{Introduction}
Neural operators provide a practical route to fast PDE surrogate modeling and have become a standard backbone for data-driven scientific simulation \citep{li2021fno,lu2021deeponet,pdebench2022}. In many downstream settings, however, raw predictive accuracy is not sufficient: decisions require uncertainty estimates that are both numerically stable and calibration-aware under distribution shift.

The dominant baselines are MC dropout, deep ensembles \citep{gal2016dropout,lakshminarayanan2017ensemble}, and post hoc conformal wrappers \citep{angelopoulos2021conformal,ma2024uqno}. While these approaches provide essential baselines, they lack a direct mechanism to optimize uncertainty against the train--test mismatch driving overfitting in neural PDE pipelines. Consequently, fixed stochastic settings often fail to adapt to regime-specific scarcity or sparsity.

We formulate uncertainty as a learnable optimization target coupled directly to the regularization objective. Concretely, we keep a predictive uncertainty branch optimized on the training split through the likelihood objective, and we introduce a separate generalization-uncertainty branch whose parameters are updated \textbf{only} using gradients computed on a regularization split. This brings the cross-regularization (XReg) principle \citep{brito2025xreg} to neural PDE surrogates, shifting the focus from architectural noise placement to objective-routed learning. Distinct from conventional hyperparameter tuning, XReg updates uncertainty parameters online using the regularization set without re-fitting the predictor backbone. Throughout, we use ``predictive'' and ``generalization'' uncertainty as operational labels defined by this gradient routing rather than as a physical decomposition.

Our contributions are:
\begin{itemize}[leftmargin=*]
    \item \textbf{Cross-regularized uncertainty learning for neural PDE surrogates.} We partition parameters into predictor and uncertainty controls and train them with a train/reg split: predictor parameters are updated on the training split, while generalization-uncertainty controls are updated online from regularization-split gradients (Algorithm~\ref{alg:xreg_dual_noise}).
    \item \textbf{Regime-adaptive uncertainty scaling from held-out gradients.} The regularization update directly adjusts uncertainty magnitude in response to train--test mismatch estimated on the regularization split, yielding uncertainty that increases when observations are sparse or data are scarce and decreases when supervision is informative.
    \item \textbf{Empirical calibration and localization evidence.} On APEBench sweeps over observed fraction and training-set size, the learned predictive distributions improve held-out Gaussian-mixture calibration relative to MC dropout and 3-member ensembles under matched protocols, and one-step spatial diagnostics show uncertainty concentrating in high-error regions.
\end{itemize}

\section{Related Work}
\textbf{Neural operator surrogates for PDEs.}
Operator-learning architectures such as DeepONet and Fourier Neural Operators are now standard baselines for PDE surrogate modeling \citep{lu2021deeponet,li2021fno}. Benchmark efforts have further clarified both their strengths and their calibration/generalization challenges across tasks and resolutions \citep{pdebench2022}.

\textbf{Uncertainty baselines in deep models.}
Widely used uncertainty baselines include MC dropout and deep ensembles \citep{gal2016dropout,lakshminarayanan2017ensemble}. Conformal approaches provide finite-sample calibration guarantees and have recently been adapted to operator-learning settings \citep{angelopoulos2021conformal,ma2024uqno}. Our focus is an adaptive coupling between train and reg.\ set objectives that tunes uncertainty scale to regime-specific mismatch.

\textbf{Validation-gradient complexity control.}
Prior work on cross-regularization (XReg) introduces learning regularization controls directly from held-out gradients \citep{brito2025xreg}. We use this principle with neural operators to disentangle predictive and generalization uncertainty within a single optimization loop.

\section{XReg Primer}
Cross-regularization (XReg) \citep{brito2025xreg} separates feature fitting from complexity control by routing their gradients through different data partitions. We decompose uncertainty into two operational categories: \emph{predictive uncertainty}, which absorbs residuals required for training fit, and \emph{generalization uncertainty}, which smooths the model to reduce train--held-out mismatch. This decomposition is model-relative rather than mechanistic: predictive uncertainty mirrors aleatoric uncertainty in absorbing residual variability (including approximation and optimization error), while generalization uncertainty captures epistemic ambiguity driven by limited evidence. The distinction between predictive and generalization uncertainty arises solely from the optimization signal rather than distinct probabilistic architectures.

\subsection{Cross-Regularized Optimization Split}
Let model parameters be partitioned into deterministic backbone parameters \(\theta\), predictive-noise parameters \(\psi\), and generalization-noise parameters \(\rho\). XReg uses alternating updates:
\begin{align}
(\theta,\psi)_{t+1} &= (\theta,\psi)_t - \eta_{\theta,\psi}\nabla_{(\theta,\psi)}\mathcal{L}_{\text{train}}((\theta,\psi)_t,\rho_t), \\
\rho_{t+1} &= \rho_t - \eta_{\rho}\nabla_{\rho}\mathcal{L}_{\text{reg}}((\theta,\psi)_{t+1},\rho_t).
\end{align}

\section{Cross-Regularized Uncertainty Framework}
\subsection{Backbone-Agnostic Formulation}
Consider a generic neural operator backbone \(f_\theta\). We augment it with two stochastic parameter sets:
\[
\psi \in \Psi \quad \text{(predictive-noise parameters)}, \qquad
\rho \in \mathcal{R} \quad \text{(generalization-noise parameters)}.
\]
For input \(x\), the model produces predictive moments
\[
\mu_{\theta,\psi,\rho}(x), \qquad \sigma_{\text{pred},\psi}(x),
\]
where \(\rho\) may affect the output distribution through stochastic perturbations of internal features and/or output layers.

\subsection{Where Noise Can Be Injected}
The framework is defined by the train/regularization optimization split, so the uncertainty pathways can be attached to many model sites. In this paper we focus on output-head and internal-feature instantiations, with the option to co-locate predictive and generalization noise while keeping them distinct through gradient routing.

For the internal variants used here, stochasticity is implemented as multiplicative Gaussian perturbations of selected latent feature blocks and, when enabled, selected Fourier-mode channels, using \(h \mapsto h\odot(1+\sigma\epsilon)\). This placement flexibility does not require a Bayesian weight posterior and avoids committing to a single architectural proxy for aleatoric versus epistemic uncertainty.

\subsection{Train and Regularization Objectives}
\textbf{Train objective.}
Train updates optimize a Monte Carlo approximation of the marginal (mixture) likelihood under generalization sampling:
\[
\mathcal{L}_{\text{train}}
= -\log \frac{1}{S}\sum_{s=1}^{S} p_{\text{hier}}(y\mid x,\omega_s),
\]
with
\[
p_{\text{hier}}
= \exp\!\left(
-\frac{1}{2}\left[\log(2\pi \sigma_{\text{pred}}^2) + \frac{(y-\mu)^2+\sigma_{\text{gen}}^2}{\sigma_{\text{pred}}^2}\right]
\right).
\]
In internal-noise variants, \(\sigma_{\text{gen}}\) is implicit in sampled model instances, and the generalization contribution is measured through \(\mathrm{Var}_{\omega}[\mu]\) at evaluation time.
Train updates optimize \((\theta,\psi)\); regularization parameters \(\rho\) are optimized only by \(\mathcal{L}_{\text{reg}}\).

\textbf{Regularization objective.}
Regularization updates are performed on held-out one-step pairs. We use two estimators in this work:
\begin{enumerate}[leftmargin=*]
    \item \textbf{Mixture NLL}:
    \[
    \mathcal{L}_{\text{reg}}^{\text{mix}}
    = -\log \frac{1}{S}\sum_{s=1}^{S} p(y\mid \mu_s, \sigma_s^2),
    \]
    where each Monte Carlo sample induces one Gaussian.
    \item \textbf{Moment-matched NLL}:
    \[
    \mu_{\text{mm}}=\mathbb{E}[\mu_s], \quad
    \sigma_{\text{mm}}^2=\mathbb{E}[\sigma_s^2+\mu_s^2]-\mu_{\text{mm}}^2,
    \]
    \[
    \mathcal{L}_{\text{reg}}^{\text{mm}}=-\log p(y\mid \mu_{\text{mm}},\sigma_{\text{mm}}^2).
    \]
\end{enumerate}

\textbf{Head-only special case (predictive and generalization branches at the same model position).}
When both branches are output heads (without internal-layer stochastic sampling), the asymmetry is explicit:
\[
\mathcal{L}_{\text{train}}
=
\frac{1}{2}\left[
\log(2\pi\sigma_{\text{pred}}^2)
\;+\;
\frac{(y-\mu)^2+\sigma_{\text{gen}}^2}{\sigma_{\text{pred}}^2}
\right],
\]
whereas the regularization objective uses total variance
\[
\sigma_{\text{tot}}^2=\sigma_{\text{pred}}^2+\sigma_{\text{gen}}^2,\qquad
\mathcal{L}_{\text{reg}}=
\frac{1}{2}\left[
\log(2\pi\sigma_{\text{tot}}^2)+\frac{(y-\mu)^2}{\sigma_{\text{tot}}^2}
\right]
\]
(or its MC-mixture counterpart when internal stochastic noise is enabled). This objective/data-routing asymmetry enables distinct learning dynamics even when predictive and generalization parameters are co-located at the output head.

\subsection{Algorithm}
\begin{figure}[t]
    \centering
    \fbox{
    \begin{minipage}{0.96\linewidth}
    \textbf{Algorithm 1: Cross-Regularized Dual-Noise Training}

    \textbf{Input:} \(\mathcal{D}_{\text{train}}, \mathcal{D}_{\text{reg}}, T, k_{\text{reg}}, \eta_{\theta,\psi}, \eta_{\rho}, S\). \\
    \textbf{Parameters:} backbone \(\theta\), predictive-noise parameters \(\psi\), generalization-noise parameters \(\rho\). \\
    \textbf{Output:} trained \((\theta,\psi,\rho)\).

    \begin{enumerate}[leftmargin=*,label=\textbf{Step \arabic*:}]
        \item Initialize \((\theta,\psi,\rho)\).
        \item For \(t=1,\dots,T\), repeat:
        \begin{enumerate}[leftmargin=1.8em,label=(\alph*)]
            \item Sample train mini-batch \(B_{\text{train}}\sim\mathcal{D}_{\text{train}}\).
            \item Sample model instances \(\omega_{1:S}\sim q_{\rho}\) and compute \(\mathcal{L}_{\text{train}}(B_{\text{train}};\theta,\psi,\omega_{1:S})\).
            \item Update
            \[
            (\theta,\psi)\leftarrow(\theta,\psi)-\eta_{\theta,\psi}\nabla_{(\theta,\psi)}\mathcal{L}_{\text{train}}.
            \]
            \item If \(t \bmod k_{\text{reg}}=0\):
            \begin{enumerate}[leftmargin=1.8em,label=(\roman*)]
                \item Sample regularization mini-batch \(B_{\text{reg}}\sim\mathcal{D}_{\text{reg}}\).
                \item Sample \(\omega_{1:S}\sim q_{\rho}\) and evaluate \(\mathcal{L}_{\text{reg}}(B_{\text{reg}};\theta,\psi,\rho,\omega_{1:S})\).
                \item Update
                \[
                \rho\leftarrow\rho-\eta_{\rho}\nabla_{\rho}\mathcal{L}_{\text{reg}}.
                \]
            \end{enumerate}
        \end{enumerate}
        \item Return \((\theta,\psi,\rho)\).
    \end{enumerate}
    \end{minipage}
    }
    \caption{Cross-regularized dual-noise training. Train updates optimize \((\theta,\psi)\) on \(\mathcal{D}_{\text{train}}\), and periodic regularization updates optimize \(\rho\) on \(\mathcal{D}_{\text{reg}}\).}
    \label{alg:xreg_dual_noise}
\end{figure}

Every iteration updates \((\theta,\psi)\) on \(\mathcal{D}_{\text{train}}\); every \(k_{\text{reg}}\) steps, a second pass updates only \(\rho\) on \(\mathcal{D}_{\text{reg}}\). Here \(q_{\rho}\) denotes the stochastic model family induced by \(\rho\), and \(\omega\sim q_{\rho}\) denotes one sampled model instance. In head-only variants, \(\omega\) corresponds to sampled output-level generalization-noise realizations; in internal-noise variants, \(\omega\) corresponds to multiplicative perturbations in selected latent feature blocks. In our main experiments, \(k_{\text{reg}}=5\); practical overhead and its dependence on \(k_{\text{reg}}\) are summarized in Appendix~\ref{app:compute_overhead}, and less frequent regularization updates approach near-standard training cost \citep{brito2025xreg}.

\subsection{Instantiation in This Work}
For empirical study, we instantiate the framework with a 1D Fourier Neural Operator backbone \citep{li2021fno}. The model outputs
\[
\mu_\theta(x_t), \quad \log \sigma_{\text{pred},\psi}(x_t), \quad \log \sigma_{\text{gen},\rho}(x_t),
\]
with output-head and internal-layer variants used across ablations and sweeps.

\subsection{Uncertainty Decomposition at Evaluation}
At inference, we separate uncertainty into predictive and generalization terms:
\begin{align}
U_{\text{pred}}(x) &= \sigma_{\text{pred}}^2(x), \\
U_{\text{gen}}(x) &= \mathrm{Var}_{\omega\sim q_\rho}\left[\mu(x;\omega)\right], \\
U_{\text{tot}}(x) &\approx U_{\text{pred}}(x) + U_{\text{gen}}(x),
\end{align}
where \(U_{\text{gen}}\) has two implementation cases used in this work:
\[
\text{head-only: }U_{\text{gen}}(x)=\sigma_{\text{gen}}^2(x),\qquad
\text{internal stochastic: }U_{\text{gen}}(x)=\mathrm{Var}_{\omega}\!\left[\mu(x;\omega)\right].
\]
We report one-step likelihood and Gaussian-mixture calibration as the primary quantitative metrics, plus spatial localization diagnostics.

\section{Experimental Setup}
\subsection{Data and Pair Construction}
The primary experiments use one-step teacher-forced pairs from APEBench trajectories, and the car experiment uses the OTNO pipeline on car-surface pressure data. Appendix~\ref{app:data_repro} documents dataset provenance, split construction, masking protocol, and metric definitions.

\subsection{Experimental Configurations Used in Figures}
Main-paper figures compare four controlled regimes:
\begin{itemize}[leftmargin=*]
    \item \textbf{Low-observation XReg regime:} output-head predictive and generalization scales under partial observations.
    \item \textbf{Head-only decomposition ablation:} both uncertainty branches at the output head, used to test whether objective/data routing alone separates their roles.
    \item \textbf{Baseline without learned regularization:} same backbone family with no active learned generalization branch.
    \item \textbf{Internal-noise XReg reference regime:} internal generalization-noise injection enabled, predictive head enabled, head-level generalization branch disabled, and moment-matched regularization.
\end{itemize}
These regimes test decomposition identifiability, calibration behavior, and the effect of where uncertainty is injected.

\section{Empirical Calibration and Dynamics}
\subsection{Mixture Calibration Comparison}
\begin{figure}[t]
    \centering
    \includegraphics[width=\linewidth]{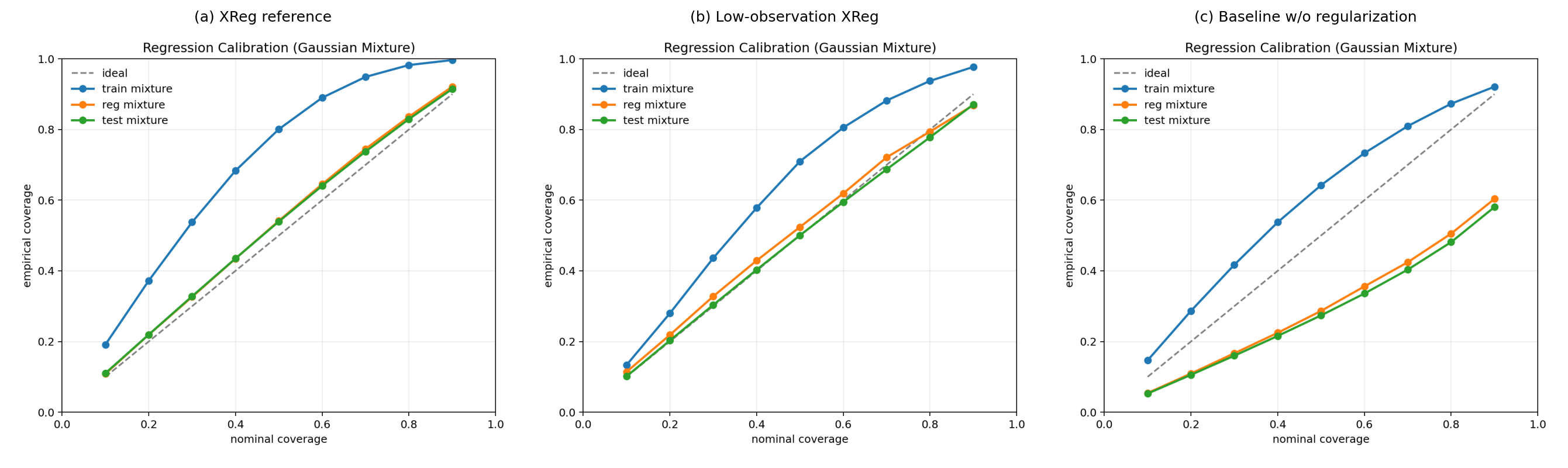}
    \caption{
    Regression-mixture calibration in three representative regimes.
    Left: XReg reference regime with moment-matched regularization.
    Middle: low-observation XReg regime (\(30\%\) observed).
    Right: model without learned regularization.
    In all panels, XReg tracks the diagonal more closely on held-out splits, with the largest gains under limited observations.
    }
    \label{fig:calibration_triplet}
\end{figure}

We prioritize calibration analysis to validate the method's ability to capture predictive distributions beyond simple mean error. Figure~\ref{fig:calibration_triplet} shows representative regimes under a matched protocol; formal metric definitions (including \(\mathrm{ECE}_{\mathrm{mix}}\) and one-step NLL normalization) are provided in Appendix~\ref{app:eval_metrics}.

\subsection{Optimization Dynamics and Learned Uncertainty Scales}
\begin{figure}[t]
    \centering
    \begin{subfigure}[t]{0.49\linewidth}
        \centering
        \includegraphics[width=\linewidth]{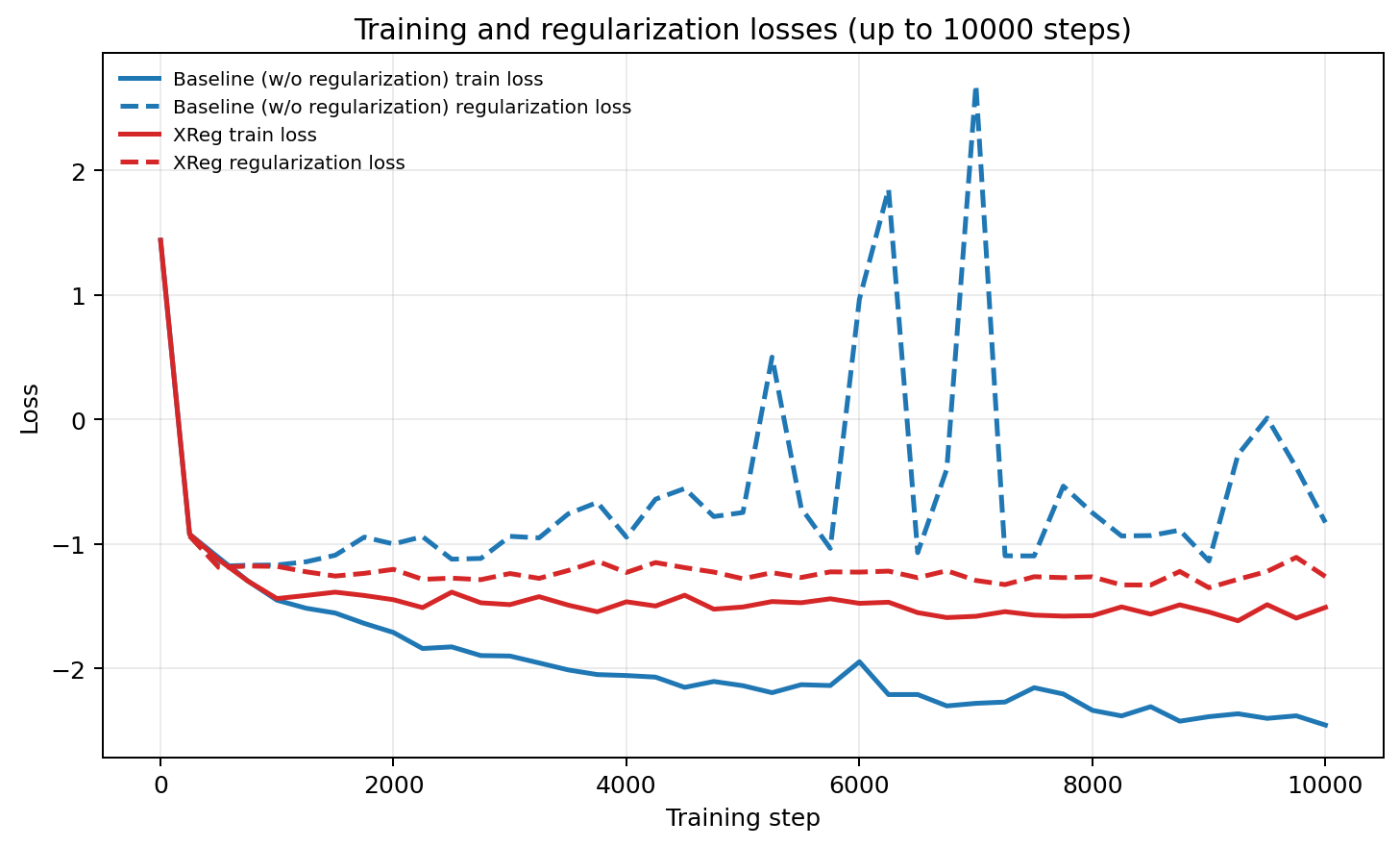}
        \caption{Train and regularization losses up to step \(10{,}000\).}
    \end{subfigure}
    \hfill
    \begin{subfigure}[t]{0.49\linewidth}
        \centering
        \includegraphics[width=\linewidth]{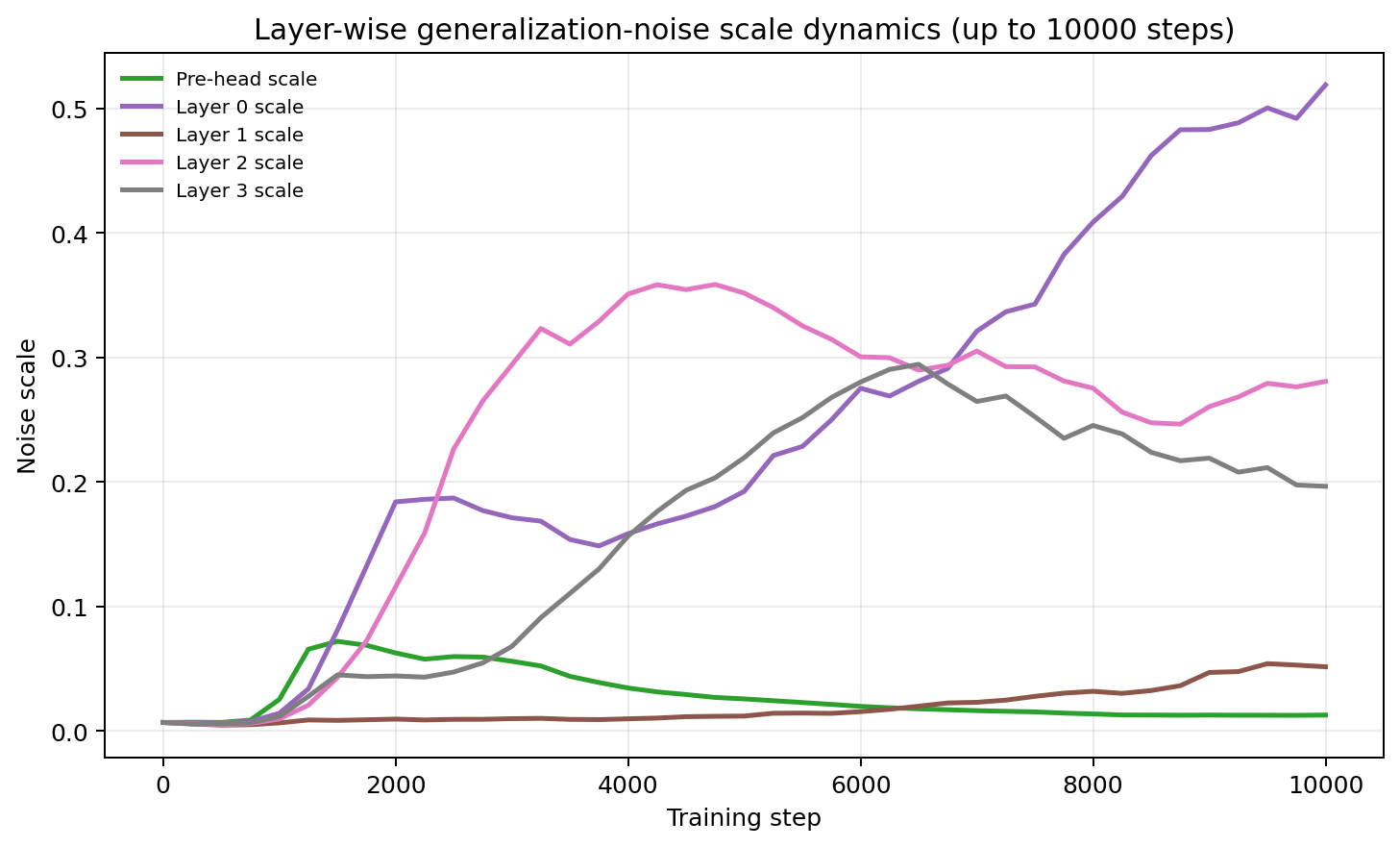}
        \caption{Layerwise evolution of generalization-noise scales up to step \(10{,}000\).}
    \end{subfigure}
    \caption{
    Optimization and uncertainty-allocation dynamics.
    Left: model without learned regularization versus XReg train/reg losses (solid=train, dashed=reg). XReg maintains a substantially tighter train-to-reg.\ set gap.
    Right: pre-head and layerwise generalization-noise scales. Uncertainty is allocated non-uniformly across layers, indicating learned structure.
    }
    \label{fig:loss_sigma_10k}
\end{figure}

Optimization trajectories (Figure~\ref{fig:loss_sigma_10k}) illustrate the coupling between train and regularization objectives and how uncertainty magnitude is distributed across injection sites.

\subsection{Spatially Resolved One-Step Uncertainty Tracking}
To test spatial selectivity of uncertainty assignment, we analyze one-step spatiotemporal maps.
\begin{figure}[t]
    \centering
    \includegraphics[width=\linewidth]{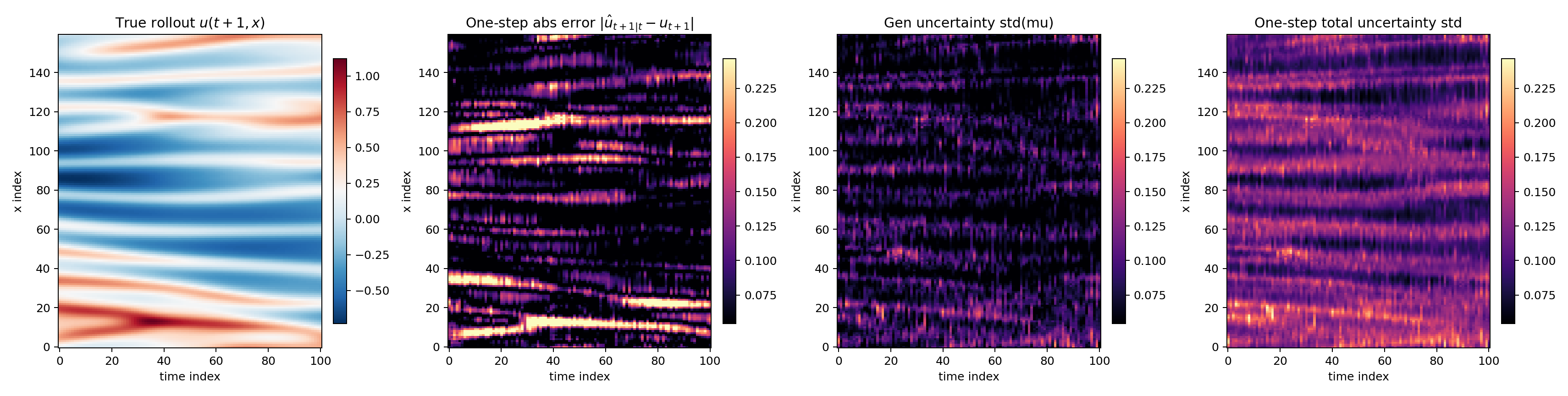}
    \caption{
    One-step teacher-forced maps (time indices \(0\) to \(100\)): true field, absolute one-step error, generalization uncertainty (\(\mathrm{Std}_{\omega}[\mu]\)), and total predictive uncertainty.
    Error and uncertainty panels share the same numeric range (set by total-uncertainty limits) to enable direct spatial comparison.
    High-error regions co-localize with high predicted uncertainty, indicating spatially selective uncertainty assignment.
    }
    \label{fig:one_step_localization}
\end{figure}

\noindent
Figure~\ref{fig:one_step_localization} compares one-step error and predicted uncertainty at matched space--time coordinates. The uncertainty field varies spatially, concentrating in regions with larger residuals.

\subsection{OTNO Car Surface-Pressure Experiment}
Transfer beyond 1D trajectories is evaluated with a geometry-conditioned car-pressure experiment using an Optimal Transport Neural Operator (OTNO) backbone \citep{li2025geometric_ot}. The data source and preprocessing provenance are documented in Appendix~\ref{app:car_data}.

The run shown in Figure~\ref{fig:car_otno_long5x} uses a deliberately data-limited split with \(n_{\text{train}}=30\) and \(n_{\text{reg}}=50\), with the remaining settings following the default protocol in Appendix~\ref{app:defaults}. Predictive uncertainty is parameterized at the output head, while generalization uncertainty is injected in internal layers.

Figure~\ref{fig:car_otno_long5x} shows three complementary views: absolute error on the car surface (left), cross-regularized uncertainty on the same surface (middle), and a roofline slice with mean and uncertainty bands (right). The spatial correspondence between the left and middle panels indicates that uncertainty concentrates in high-error regions and remains spatially selective. The slice view shows contributions from both predictive and generalization uncertainty, with uncertainty envelopes widening around difficult regions and narrowing where the fit is stable. Figure~\ref{fig:appendix_car_sigma_long5x} reports additional uncertainty-scale trajectories for this run.

\begin{figure}[t]
    \centering
    \includegraphics[width=\linewidth]{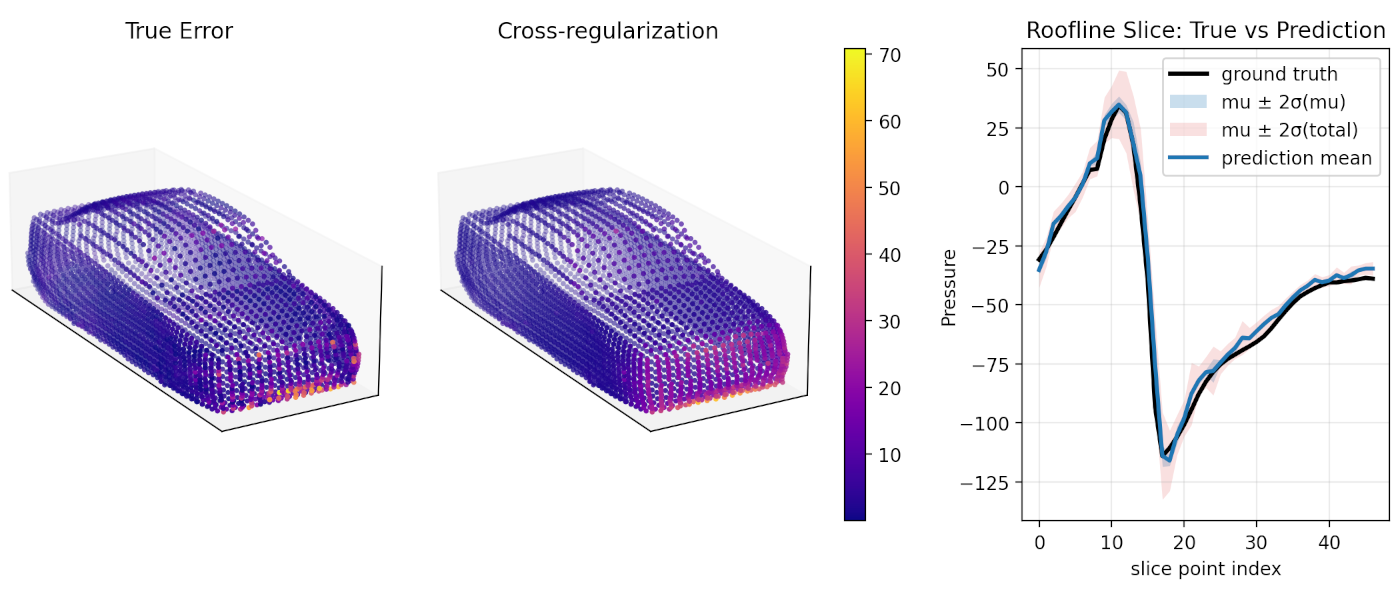}
    \caption{
    OTNO car-pressure experiment with cross-regularized uncertainty (\(n_{\text{train}}=30,\;n_{\text{reg}}=50\)).
    Left: absolute prediction error on the car surface.
    Middle: learned uncertainty field on the same sample.
    Right: roofline slice with ground truth, predictive mean, and uncertainty bands.
    Error hot spots and uncertainty hot spots are spatially aligned, and the slice shows calibrated widening of uncertainty in regions with larger residuals.
    }
    \label{fig:car_otno_long5x}
\end{figure}

\subsection{Sweep Comparison: XReg vs.\ Baselines (Loss + ECE)}
\begin{figure}[t]
    \centering
    \includegraphics[width=0.82\linewidth]{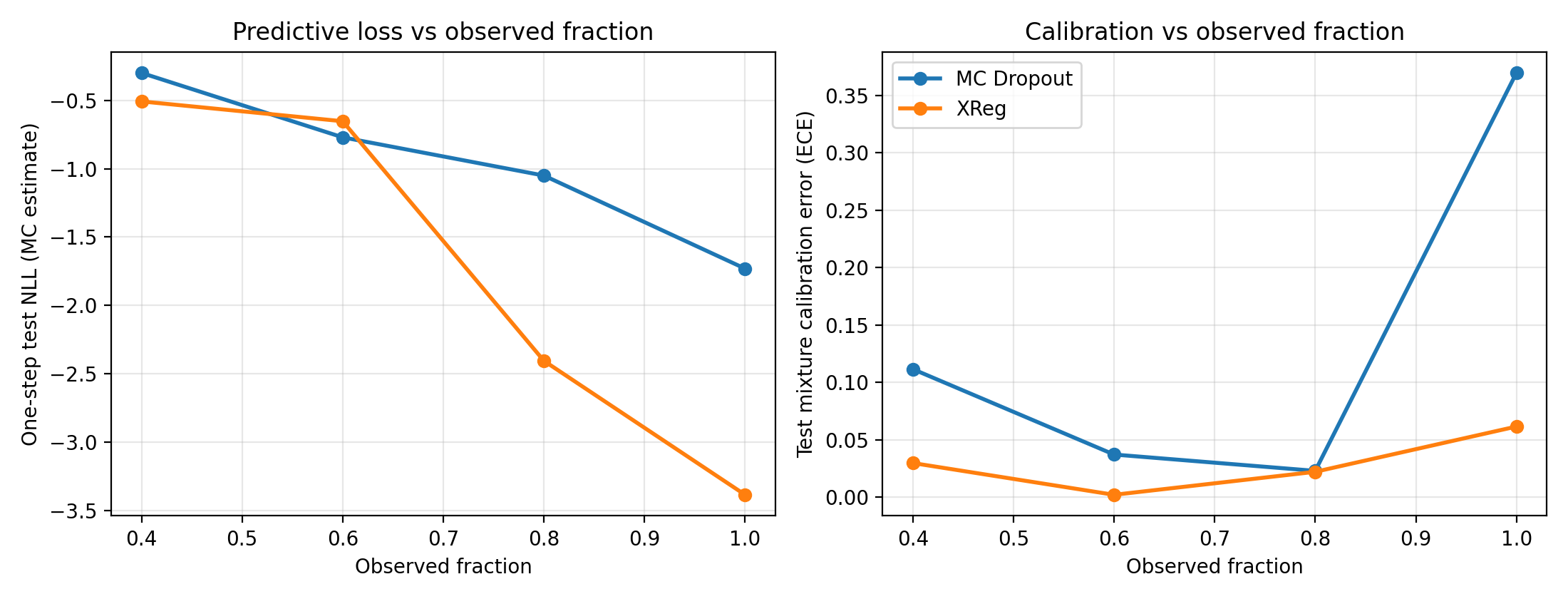}
    \caption{
    Observed-fraction sweep (\(0.4,0.6,0.8,1.0\)) comparing XReg and MC dropout on one-step test NLL (left, MC-evaluated) and test mixture ECE (right).
    The two-panel view separates fit quality from calibration quality.
    XReg is consistently better calibrated (lower ECE) across all observation fractions while preserving competitive one-step NLL.
    }
    \label{fig:xreg_vs_mcdrop_obs_curve}
\end{figure}

XReg is compared against two stochastic baselines under matched observation settings: MC dropout (\(p=0.1\)) and deep ensembles (3 members). Dropout \(p=0.1\) is used as a compute-matched baseline; tuning \(p\) per regime is possible but would undermine the no-hand-tuning objective of this comparison. Figure~\ref{fig:xreg_vs_mcdrop_obs_curve} compares performance scaling between XReg and MC dropout. Exact observed-fraction values are reported in Table~\ref{tab:xreg_vs_mcdrop_obs}, and the train-size sweep values are reported in Appendix Table~\ref{tab:xreg_vs_mcdrop_train_size}. Deep-ensemble values are computed from the aggregated predictive mixture of the three trained members at each sweep point.

\begin{table}[t]
\centering
\small
\setlength{\tabcolsep}{4pt}
\begin{tabular}{llrrr}
\toprule
Obs. Frac & Method & Test NLL (MC)$\downarrow$ & Test ECE$_{\text{mix}}\downarrow$ & Reg ECE$_{\text{mix}}\downarrow$ \\
\midrule
0.4 & XReg & \textbf{-0.5083} & \textbf{0.0296} & \textbf{0.0248} \\
0.4 & MC Dropout & -0.2999 & 0.1115 & 0.1153 \\
0.4 & Deep Ensemble (3) & -0.1002 & 0.1121 & 0.1125 \\
\midrule
0.6 & XReg & -0.6533 & \textbf{0.0020} & \textbf{0.0040} \\
0.6 & MC Dropout & -0.7710 & 0.0370 & 0.0324 \\
0.6 & Deep Ensemble (3) & \textbf{-0.8900} & 0.0491 & 0.0522 \\
\midrule
0.8 & XReg & \textbf{-2.4052} & \textbf{0.0220} & \textbf{0.0139} \\
0.8 & MC Dropout & -1.0506 & 0.0228 & 0.0159 \\
0.8 & Deep Ensemble (3) & -2.0772 & 0.0624 & 0.0530 \\
\midrule
1.0 & XReg & -3.3849 & \textbf{0.0614} & \textbf{0.0596} \\
1.0 & MC Dropout & -1.7310 & 0.3696 & 0.3695 \\
1.0 & Deep Ensemble (3) & \textbf{-3.5632} & 0.2943 & 0.2972 \\
\bottomrule
\end{tabular}
\caption{
    Numeric comparison for the observed-fraction sweep.
    XReg achieves the lowest test and regularization mixture ECE across all four observation fractions, while maintaining competitive one-step likelihood.
}
\label{tab:xreg_vs_mcdrop_obs}
\end{table}

\noindent
These results demonstrate that cross-regularized uncertainty scales are learned directly from regime-specific train--test mismatch. By construction, XReg updates generalization uncertainty from regularization-split gradients, allowing the uncertainty scale to track observation level and data regime during training. Relative to fixed-rate dropout and 3-member deep ensembles, this yields more consistent calibration across data and observability conditions.

\subsection{Generalization Uncertainty Tracks Noise and Data Availability}
\begin{figure}[t]
    \centering
    \includegraphics[width=0.88\linewidth]{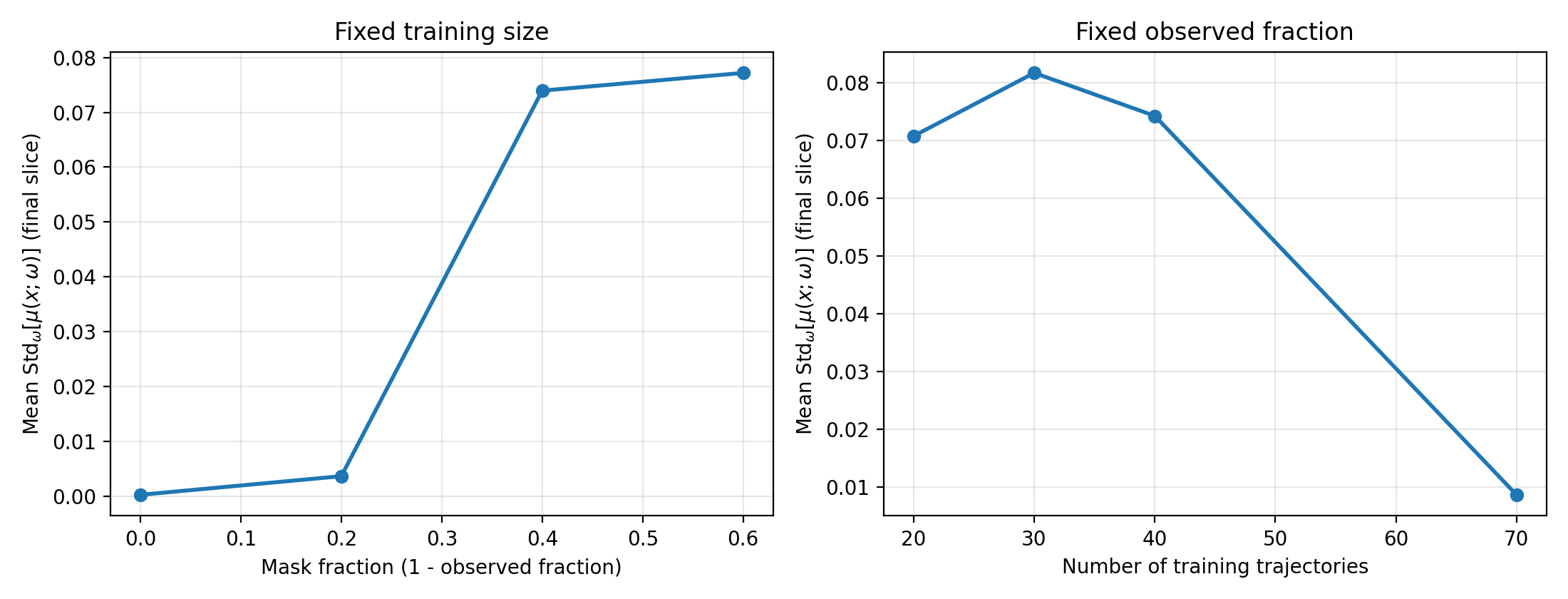}
    \caption{
    Noise/data scaling of learned generalization uncertainty, measured as mean \(\mathrm{Std}_{\omega}[\mu]\) on the final slice.
    Left: increasing mask fraction (fixed train size \(=50\)) corresponds to less observed signal and produces larger learned generalization uncertainty.
    Right: increasing train-set size (fixed observed fraction \(=0.7\)) reduces learned generalization uncertainty.
    These trends indicate that XReg adapts uncertainty magnitude to regime difficulty.
    }
    \label{fig:stdmu_combined}
\end{figure}

Figure~\ref{fig:stdmu_combined} isolates how the learned generalization uncertainty responds to regime difficulty. When the observed fraction decreases (weaker supervision), the learned generalization uncertainty increases; when the training-set size increases (more evidence), it decreases. Under the cross-regularized update, regularization-split gradients set the uncertainty scale in the direction predicted by train--test mismatch, producing regime-dependent noise levels.

\subsection{Main Findings and Implications}
Across matched protocols, cross-regularized uncertainty improves held-out mixture calibration while preserving competitive one-step likelihood, indicating that the calibration gains are not explained by trivial variance inflation. The supporting diagnostics are consistent with the proposed mechanism: optimization traces show a reduced train-to-regularization gap; learned uncertainty scales evolve non-uniformly across layers; and one-step spatial maps concentrate uncertainty in regions that exhibit larger one-step error.

\section{Discussion: Model-Relative Uncertainty}
Uncertainty in neural PDE surrogates is operational and therefore model-relative. Predictive uncertainty combines irreducible observation noise (including masking and measurement noise) with residual structure arising from approximation limits, optimization error, and data scarcity. Generalization uncertainty captures the sensitivity of the fitted model to train-specific patterns that fail to transfer to held-out data, resembling an epistemic component. Here it is learned by optimizing uncertainty controls with regularization-split gradients, without requiring a prior over weights or a posterior approximation. Appendix~\ref{app:extended_discussion} discusses implications for chaotic regimes and clarifies the distinction from conventional hyperparameter optimization.

\section{Conclusion}
We presented cross-regularized uncertainty learning for neural PDE surrogates, where uncertainty parameters are trained online using regularization-split gradients while predictor parameters are trained on the training split. This train/reg routing yields an operational separation between predictive uncertainty (fit-time residual absorption) and generalization uncertainty (train--held-out mismatch control) without ensembles or Bayesian posterior approximations. Empirically, the resulting predictive distributions improve one-step Gaussian-mixture calibration under partial observability and limited data, and spatial diagnostics show uncertainty concentrating in high-error regions. Extending these guarantees from one-step teacher-forced evaluation to calibrated long-horizon autoregressive rollouts requires propagating uncertainty through compounding model error; we leave that setting to future work.

A related direction is richer predictive noise inside the model interior. Predictive noise injected at internal layers can, in principle, induce correlated or multimodal output uncertainty. Extending the current instantiations in that direction may narrow the gap between single-model uncertainty learning and richer uncertainty families used in scientific decision-making.

\bibliographystyle{iclr2026_conference}
\bibliography{references}

\appendix
\section{Appendix: Datasets, Splits, and Reproducibility}
\label{app:data_repro}
\subsection{Default Training Configuration}
\label{app:defaults}
Unless explicitly overridden in a sweep axis, runs use:
\begin{itemize}[leftmargin=*]
    \item a 4-layer FNO backbone with 12 Fourier modes, hidden width 8, and GELU activations,
    \item batch size \(16\),
    \item total optimization steps \(30{,}000\),
    \item evaluation interval \(250\) steps,
    \item one regularization update every 5 training steps,
    \item 10 Monte Carlo samples for train, regularization, and test evaluations,
    \item learning rates \(10^{-3}\) for backbone weights, \(10^{-2}\) for internal predictive-noise scales, \(10^{-2}\) for internal generalization-noise scales, and \(10^{-3}\) for head-level generalization-noise scales,
    \item initialization \(\log \sigma_{\text{pred}}=-5\), \(\log \sigma_{\text{gen}}=-5\).
\end{itemize}

\subsection{APEBench Data and Task Construction}
All primary experiments are built from APEBench \citep{koehler2024apebench} one-dimensional trajectory data under the \texttt{diff\_ks} setting (160 spatial points). From each trajectory, we construct one-step supervised pairs \((x_t, x_{t+1})\) for teacher-forced training and evaluation. Unless stated otherwise, train horizon is \(10\) and test horizon is \(200\). We report one-step metrics directly on these pairs because the model is trained to quantify one-step uncertainty. Long-horizon rollout calibration is future work and is nontrivial: uncertainty must remain calibrated after repeated autoregressive propagation with compounding prediction error.

\subsection{OTNO Car Data Provenance}
\label{app:car_data}
The car-pressure experiment uses the ShapeNetCar slice from the DINOZAUR data release \citep{matveev2025dinozaur,dinozaur_dataset_2025}, using the local \texttt{data/shapenetcar} split prepared for these experiments. In this setup, each sample is a car-surface point set with geometry/flow-derived inputs and scalar surface pressure targets. Our OTNO model follows the neuraloperator implementation protocol, and OTNO denotes \emph{Optimal Transport Neural Operator} \citep{li2025geometric_ot}.

\subsection{Train/Reg/Test Splits}
The optimization protocol uses an explicit three-way split:
\begin{itemize}[leftmargin=*]
    \item \textbf{train split:} used to optimize deterministic model weights and predictive uncertainty parameters;
    \item \textbf{regularization set (reg.\ set):} used only for updates of generalization-noise parameters via regularization objectives;
    \item \textbf{test split:} held out for final one-step and calibration reporting.
\end{itemize}
This split is central to the method claim: predictive and generalization uncertainty are separated by data routing and objective routing, with parameter naming reflecting that separation.

\subsection{Observation Masking Protocol}
Spatial masking is controlled by the observed fraction of input and target fields. We use a controlled-occlusion protocol: for each run, masks are sampled once with a fixed seed and reused throughout training to remove mask stochasticity from method comparisons. In all sweep plots, observation settings are matched between compared methods (XReg vs.\ MC dropout), so differences are attributable to uncertainty-learning strategy with data visibility held fixed. Random-per-step masking is a relevant extension for future work.

\subsection{Evaluation Metrics and Logging Conventions}
\label{app:eval_metrics}
Mixture calibration is measured with
\[
\mathrm{ECE}_{\mathrm{mix}}=\frac{1}{|\mathcal{A}|}\sum_{\alpha\in\mathcal{A}}
\left|\widehat{\mathrm{Coverage}}(\alpha)-\alpha\right|,
\]
where \(\mathcal{A}=\{0.1,\ldots,0.9\}\). For each \(\alpha\), \(\widehat{\mathrm{Coverage}}(\alpha)\) is empirical central-interval coverage computed from the Gaussian-mixture predictive CDF and averaged over space-time one-step targets. One-step NLL is reported per spatial target point.

We report:
\begin{itemize}[leftmargin=*]
    \item one-step test NLL from MC predictive evaluation,
    \item mixture calibration error (ECE) on regularization and test splits,
    \item calibration curves over nominal coverage grid points,
    \item uncertainty decomposition statistics including \(\mathrm{Std}_{\omega}[\mu]\) and layerwise generalization-noise scales.
\end{itemize}

\subsection{Computational Overhead}
\label{app:compute_overhead}
XReg adds a periodic regularization update for \(\rho\). If one train update has cost \(C_{\text{train}}\), one regularization update has cost \(C_{\text{reg}}\), and regularization is applied every \(k_{\text{reg}}\) steps, the approximate overhead factor is
\[
\text{overhead}\approx 1+\frac{1}{k_{\text{reg}}}\frac{C_{\text{reg}}}{C_{\text{train}}}.
\]
In our default setting (\(k_{\text{reg}}=5\)), this is a moderate overhead because \(\rho\)-updates are sparse and low-dimensional. The cost can be reduced further by increasing \(k_{\text{reg}}\) (less frequent regularization updates). Prior empirical reports indicate that practical settings can approach roughly \(1.1\times\) standard training cost \citep{brito2025xreg}, comparable to ordinary training pipelines that already include periodic validation checks.

\section{Appendix: Extended Discussion on Interpretation and Optimization}
\label{app:extended_discussion}
This section expands two discussion points that are secondary to the main empirical claims.

\textbf{Model-relative uncertainty in deterministic regimes.}
The model-relative interpretation is especially relevant in domains where ``intrinsic noise'' is debated. Chaotic systems are an illustrative case: the governing dynamics may be deterministic, yet finite data and finite-capacity models still produce large predictive residuals. In this setting, predictive uncertainty tracks unexplained residual structure, while generalization uncertainty tracks overfitting to the specific training sample.

\textbf{Why adaptation remains stable.}
The adaptation mechanism learns uncertainty magnitude from observed train--held-out mismatch during optimization. In our setup, this remains stable because regularization variables are low-dimensional noise controls, not full feature maps; the regularization step adjusts uncertainty scales without re-fitting the full predictor on held-out data.

\textbf{Relation to hyperparameter optimization.}
XReg performs direct parameter learning of uncertainty controls within a single training process. Classical hyperparameter tuning optimizes external meta-parameters (e.g., a fixed \(\lambda\) selected by outer-loop search), whereas XReg updates model parameters that control uncertainty. In Algorithm~\ref{alg:xreg_dual_noise}, \((\theta,\psi)\) are updated on train data, and \(\rho\) is updated online from reg.\ set gradients.

\section{Appendix: OTNO Car Uncertainty Diagnostics}
This appendix reports optimization-time uncertainty-scale trajectories and final uncertainty magnitudes for the OTNO car experiment in Figure~\ref{fig:car_otno_long5x}. The goal is to show that both predictive and generalization branches remain active and non-degenerate in the data-limited car regime.
\begin{figure}[h]
    \centering
    \includegraphics[width=\linewidth]{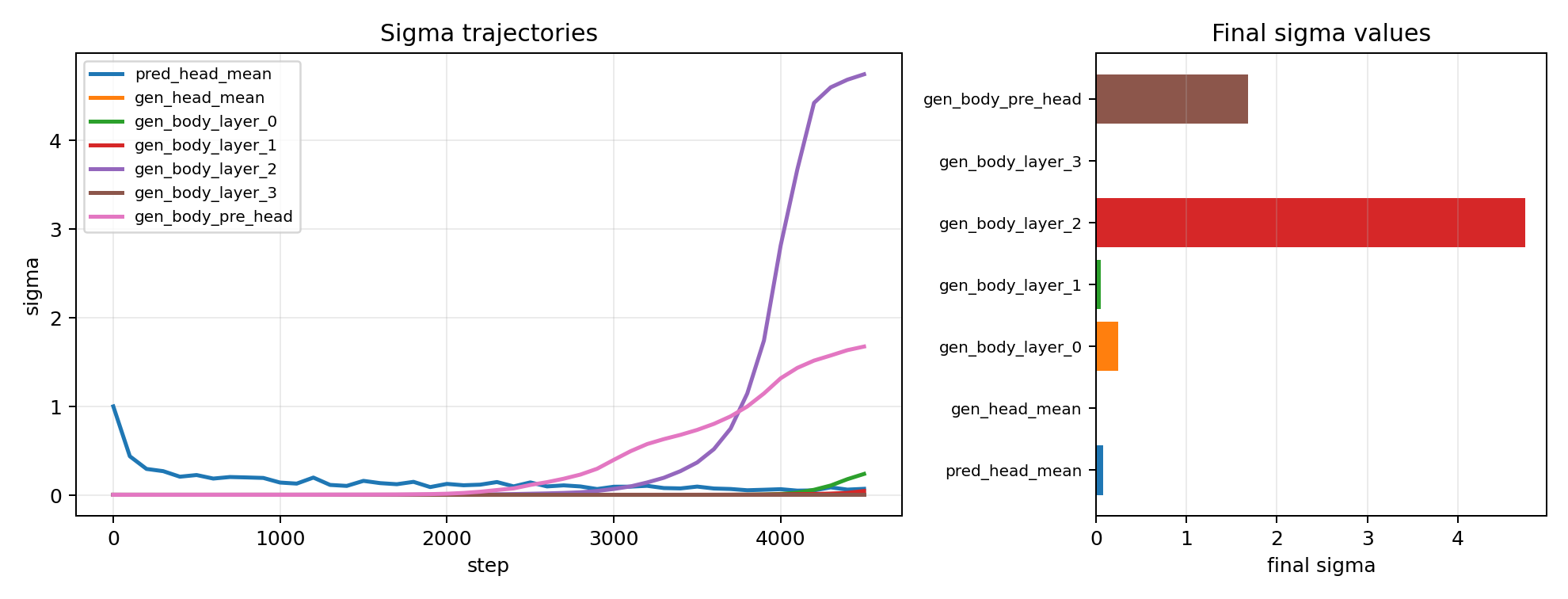}
    \caption{
    Sigma diagnostics for the car experiment shown in Figure~\ref{fig:car_otno_long5x}.
    Left: trajectory of learned uncertainty scales during optimization.
    Right: final values across uncertainty components.
    The run exhibits non-degenerate predictive and generalization scales, with clear growth in selected generalization components and no collapse to near-zero values.
    }
    \label{fig:appendix_car_sigma_long5x}
\end{figure}

\section{Appendix: Sweep Calibration Panels}
This appendix provides full calibration montages for all runs in the XReg and MC-dropout sweeps summarized in the main text tables and aggregate figures. These panels serve as a full visual audit trail beyond scalar ECE summaries.

\begin{table}[h]
\centering
\small
\setlength{\tabcolsep}{4pt}
\begin{tabular}{llrrr}
\toprule
Train Size & Method & Test NLL (MC)$\downarrow$ & Test ECE$_{\text{mix}}\downarrow$ & Reg ECE$_{\text{mix}}\downarrow$ \\
\midrule
20 & XReg & -0.3277 & \textbf{0.0965} & \textbf{0.0867} \\
20 & MC Dropout & \textbf{-0.4343} & 0.1254 & 0.1197 \\
20 & Deep Ensemble (3) & 0.7512 & 0.1344 & 0.1361 \\
\midrule
30 & XReg & -0.5962 & \textbf{0.0273} & \textbf{0.0210} \\
30 & MC Dropout & \textbf{-0.6894} & 0.0857 & 0.0815 \\
30 & Deep Ensemble (3) & 0.1574 & 0.1157 & 0.1131 \\
\midrule
40 & XReg & -0.6671 & \textbf{0.0144} & \textbf{0.0118} \\
40 & MC Dropout & \textbf{-0.8858} & 0.0375 & 0.0410 \\
40 & Deep Ensemble (3) & -0.7355 & 0.0760 & 0.0727 \\
\midrule
70 & XReg & -1.4173 & \textbf{0.0118} & \textbf{0.0128} \\
70 & MC Dropout & -1.1361 & 0.0666 & 0.0654 \\
70 & Deep Ensemble (3) & \textbf{-2.0064} & 0.0569 & 0.0555 \\
\bottomrule
\end{tabular}
\caption{
    Numeric comparison for the train-size sweep at fixed observed fraction \(0.7\).
    XReg maintains substantially lower test/reg mixture ECE across all four train sizes, including regimes where baseline one-step NLL is competitive.
}
\label{tab:xreg_vs_mcdrop_train_size}
\end{table}
\begin{figure}[h]
    \centering
    \includegraphics[width=\linewidth]{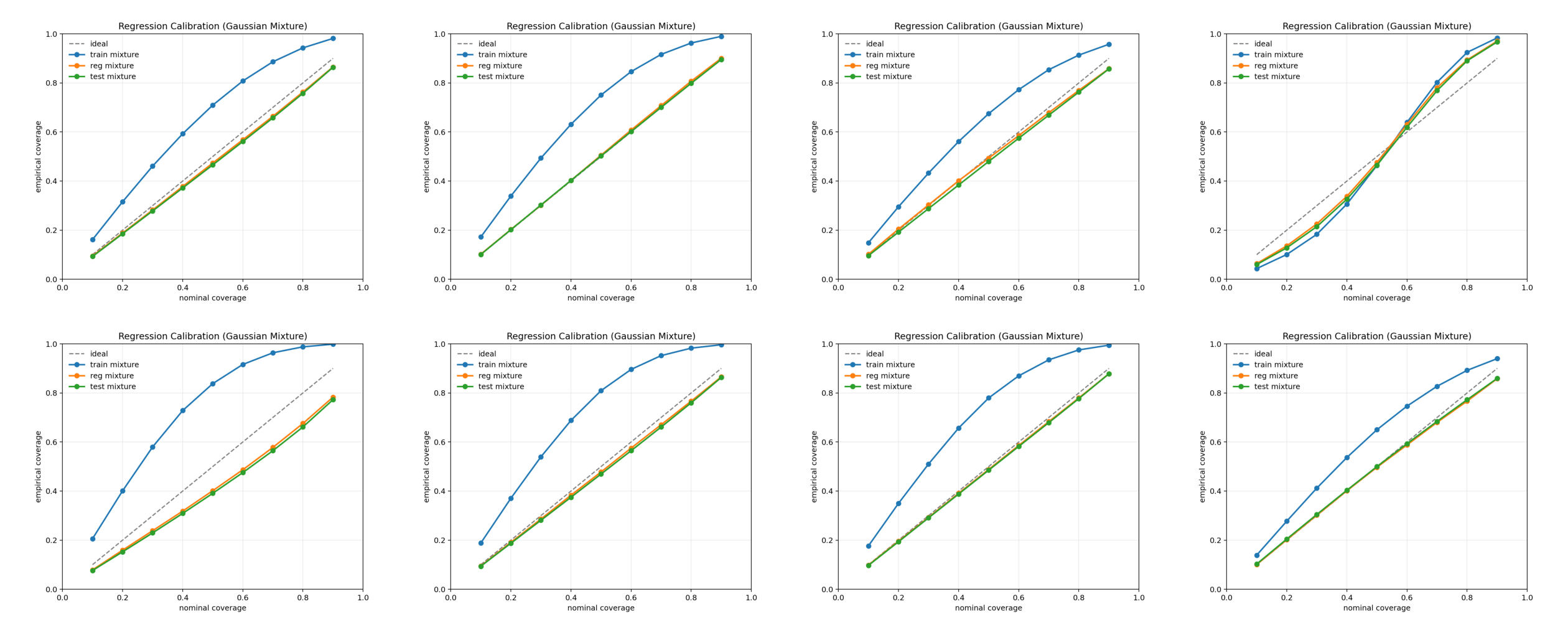}
    \caption{
    Full XReg sweep calibration montage (regression-mixture calibration for all eight XReg runs in the 10k-step sweep reported in the main text).
    These panels are included as a raw visual audit trail: they expose per-run curve shape differences that are not fully captured by scalar ECE summaries.
    }
    \label{fig:appendix_xreg_calib_montage}
\end{figure}

\begin{figure}[h]
    \centering
    \includegraphics[width=\linewidth]{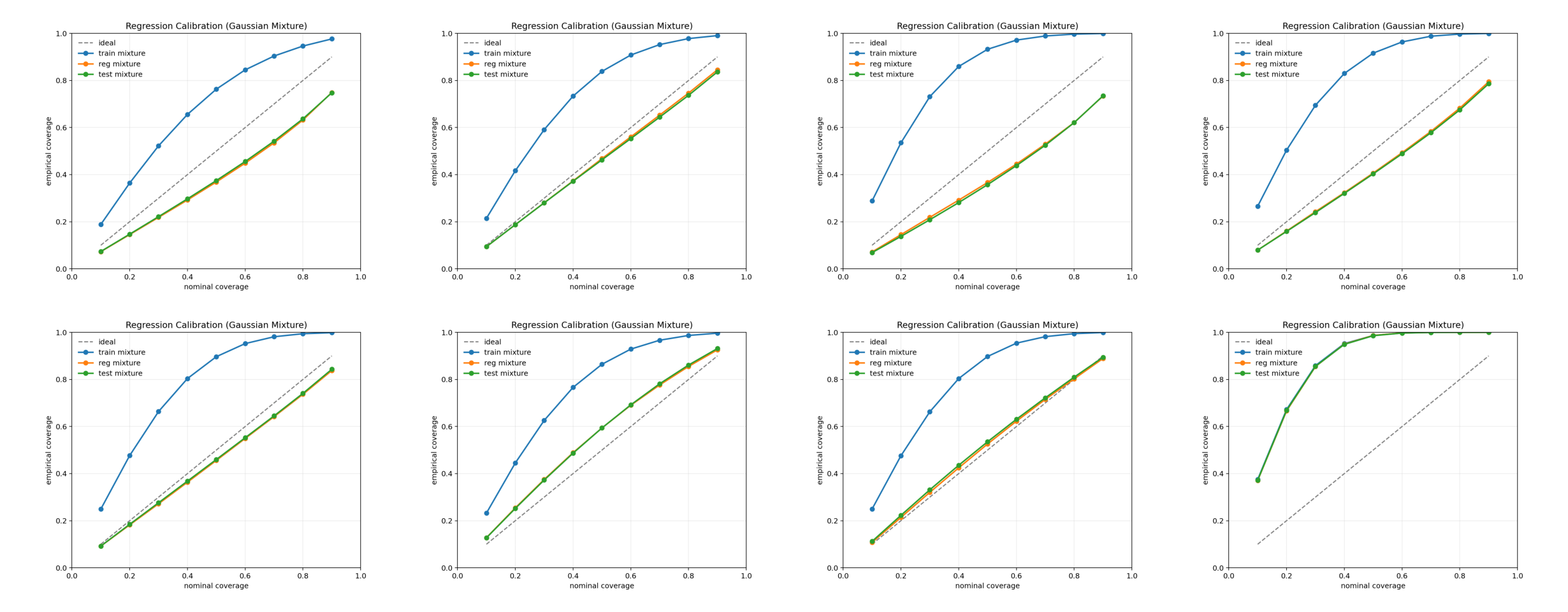}
    \caption{
    MC-dropout sweep calibration montage for the full completed 8-run sweep (observed-fraction axis and train-size axis, dropout \(p=0.1\)).
    Along with Figure~\ref{fig:appendix_xreg_calib_montage}, these panels provide a many-panel calibration comparison complementing the aggregate loss/ECE table in the main text.
    }
    \label{fig:appendix_mcdrop_calib_montage}
\end{figure}

\begin{figure}[h]
    \centering
    \includegraphics[width=\linewidth]{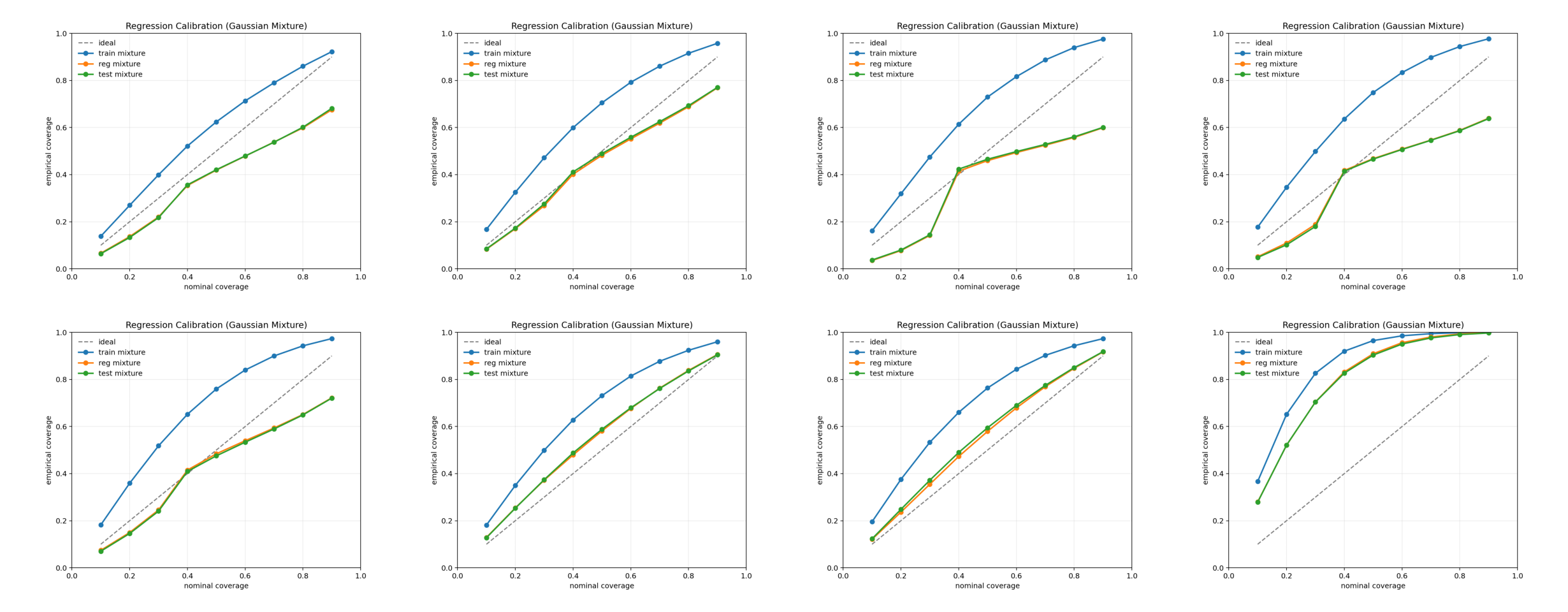}
    \caption{
    Deep-ensemble sweep calibration montage for the full completed 8-point sweep (observed-fraction axis and train-size axis), with each point evaluated as a 3-member aggregated predictive mixture.
    This view complements Figures~\ref{fig:appendix_xreg_calib_montage} and~\ref{fig:appendix_mcdrop_calib_montage} with a full-panel visual audit for the ensemble baseline.
    }
    \label{fig:appendix_deepens_calib_montage}
\end{figure}

\end{document}

%% file: math_commands.tex
\usepackage{amsmath,amsfonts,bm}

\def\eqref#1{equation~\ref{#1}}

\def\1{\bm{1}}

\DeclareMathAlphabet{\mathsfit}{\encodingdefault}{\sfdefault}{m}{sl}
\SetMathAlphabet{\mathsfit}{bold}{\encodingdefault}{\sfdefault}{bx}{n}

